\pdfoutput=1

\documentclass[11pt]{article}

\usepackage[]{EMNLP2023}

\usepackage{times}
\usepackage{latexsym}
\usepackage{mdframed}
\usepackage[T1]{fontenc}

\usepackage[utf8]{inputenc}

\usepackage{microtype}

\usepackage{inconsolata}

\usepackage{caption}
\usepackage{subcaption}
\usepackage{enumitem}
\usepackage{flushend}
\usepackage{balance}
\usepackage{lineno}
\usepackage{amsmath,amssymb,amsfonts}
\usepackage{algorithm}
\usepackage{algorithmic}
\usepackage{graphicx}
\usepackage{textcomp}
\usepackage{xcolor}
\usepackage{colortbl}
\usepackage{amsthm}
\usepackage{url}
\usepackage{float}
\usepackage{multirow}
\usepackage{multicol}
\usepackage{color}
\usepackage{bm}
\usepackage{bbm}

\usepackage{pifont}
\usepackage{array}
\newcolumntype{L}[1]{>{\raggedright\let\newline\\\arraybackslash\hspace{0pt}}m{#1}}
\newcolumntype{C}[1]{>{\centering\let\newline  \\\arraybackslash\hspace{0pt}}m{#1}}
\newcolumntype{R}[1]{>{\raggedleft\let\newline \\\arraybackslash\hspace{0pt}}m{#1}}

\usepackage{ bbold }

\DeclareMathOperator*{\argmin}{argmin}

\usepackage[utf8]{inputenc} 
\usepackage[T1]{fontenc}    
\usepackage{hyperref}       
\usepackage{url}            
\usepackage{booktabs}       
\usepackage{amsfonts}       
\usepackage{nicefrac}       
\usepackage{microtype}      
\usepackage{xcolor}         

\title{Interpreting Pretrained Language Models via Concept Bottlenecks}

\author{Zhen Tan \\
Arizona State University\\
  \texttt{ztan36@asu.edu}\vspace{0.15in} \\
     {\bf Yuan Bo} \\
Zhejiang University\\
  \texttt{byuan@zju.edu.cn} \\
  \And
  Lu Cheng \\
  University of Illinois Chicago\\
  \texttt{lucheng@uic.edu}\vspace{0.15in} \\ 
    {\bf  Jundong Li}\\
  University of Virginia\\
   \texttt{jundong@virginia.edu} 
   \And
  Song Wang \\
  University of Virginia\\
  \texttt{sw3wv@virginia.edu}\vspace{0.15in} \\ 
    {\bf  Huan Liu}\\
  Arizona State University\\
   \texttt{huanliu@asu.edu}
  }
  
\begin{document}
\maketitle
\begin{abstract}

Pretrained language models (PLMs) have made significant strides in various natural language processing tasks. However, the lack of interpretability due to their ``black-box'' nature poses challenges for responsible implementation. Although previous studies have attempted to improve interpretability by using, e.g., attention weights in self-attention layers, these weights often lack clarity, readability, and intuitiveness. In this research, we propose a novel approach to interpreting PLMs by employing high-level, meaningful concepts that are easily understandable for humans. For example, we learn the concept of ``Food'' and investigate how it influences the prediction of a model's sentiment towards a restaurant review. We introduce C$^3$M, which combines human-annotated and machine-generated concepts to extract hidden neurons designed to encapsulate semantically meaningful and task-specific concepts. Through empirical evaluations on real-world datasets, we manifest that our approach offers valuable insights to interpret PLM behavior, helps diagnose model failures, and enhances model robustness amidst noisy concept labels.
\end{abstract}

\section{Introduction}
\begin{figure}[t]
\vspace{-0.1cm}
		\centering
\scalebox{0.47}{
\includegraphics[width=0.8\textwidth]{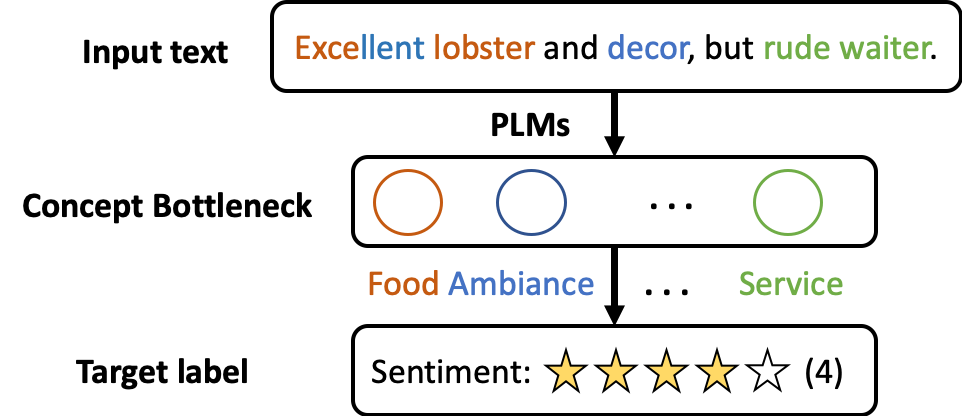}}
\vspace{-0.1cm}
\caption{The illustration of CBE-PLMs. Via PLMs, the original texts $x$ is first map into an intermediate layer consisting of a set of human-comprehensible concepts $c$, which are then used to predict the target label $y$.}
\vspace{-0.6cm}
\label{fig:example}
	\end{figure}

Although Pretrained Language Models (PLMs) like BERT~\citep{devlin2018bert} have achieved remarkable success in various NLP tasks \citep{zhuincorporating,liu2019text}, they are frequently regarded as black boxes, posing significant obstacles to their responsible deployment in real-world scenarios, particularly in critical domains such as healthcare~\cite{koh2020concept}. Therefore, enabling PLMs' interpretability is crucial to achieve socially responsible AI \cite{cheng2021socially}. To date, many existing works~\citep{belinkov2019analysis,madsen2022post} leverage attention weights extracted from the self-attention layers to provide token-level or phrase-level importance. These low-level explanations are found unfaithful \citep{yin2022interpreting} and lack readability and intuitiveness \citep{losch2019interpretability}, leading to unstable or even unreasonable explanations. To address these limitations, we seek to {explain via human-comprehensible \textit{concepts} that use more abstract features (e.g., general notions) as opposed to raw input features at the token level \citep{zarlenga2022concept,liao2023ai}. The foundation of this work is the Concept Bottleneck Models (CBMs)~\citep{koh2020concept} that interprets deep models (e.g., ResNet~\citep{he2016deep}) for image classification tasks using high-level concepts (e.g., shape).
For NLP tasks such as sentiment analysis, concepts can be Food, Ambiance, and Service as shown in Figure~\ref{fig:example}, where each concept corresponds to a neuron in the concept bottleneck layer. The final decision layer is then a linear function of these concepts. Using concepts greatly improves the readability and intuitiveness of the explanations compared to low-level features such as ``lobster''.}

\vspace{-0.1cm}
We propose to study \textit{Concept-Bottleneck-Enabled Pretrained Language Models} (CBE-PLMs). There are three key challenges: 
{First, CBMs cannot be directly adapted since PLMs are pre-trained and fine-tuned on separate corpora while CBMs work on the same end-to-end image classification tasks during training and testing. Therefore, the corpora used for pre-training PLMs may contain useful text-concept correlations that are unseen in the downstream task.
An investigation of the adaptability of CBMs to CBE-PLMs is needed. 
Second, the majority of existing CBMs \citep{koh2020concept,zarlenga2022concept} require human-annotated concepts. This can be challenging for natural language since the annotator may need to read through the entire text to understand the context and label one concept~\citep{nemeth2020machine}. This limits the practical usage and scalability of CBE-PLMs. Third, many studies have identified the tradeoff between interpretability and task accuracy using CBMs since the predetermined concepts may leave out important information for target task prediction~\citep{zarlenga2022concept}. Therefore, it is crucial to improve both interpretability and task performance to achieve optimal interpretability-utility tradeoff.}

{To tackle the first challenge, we adapt standard training strategies in CBMs~\citep{koh2020concept} to learning CBE-PLMs and conduct comprehensive analyses to identify the best way to adapt CBMs to interpret PLMs. 
For the second challenge of concept discovery and labeling, we propose leveraging Large Language Models (LLMs) trained on extensive human-generated corpora and feedbacks, such as ChatGPT~\citep{openai2023gpt4}, to identify novel concepts in text and generate pseudo-labels (via prompting) for unlabeled concepts.
Recent studies~\citep{bommasani2022opportunities,openai2023gpt4} exhibit that these LLMs encapsulate significant amounts of human common sense knowledge.
By augmenting the small set of human-specified concepts with machine-generated concepts, we increase concept diversity and useful information for prediction. In addition, generated pseudo-labels offer us a large set of instances with noisy concept labels, complementing the smaller set of instances with clean labels. To further improve interpretability-utility tradeoff (third challenge), we propose to learn from noisy concept labels and incorporate a concept-level MixUp mechanism~\citep{zhang2017mixup} that allows CBE-PLMs to cooperatively learn from both noisy and clean concept sets.} We name our framework for training CBE-PLMs as \textit{\underline{C}hatGPT-guided \underline{C}oncept augmentation with \underline{C}oncept-level \underline{M}ixup} (C$^3$M). In summary, our contributions include:
\vspace{-0.06in}
\begin{itemize}[leftmargin=*, itemsep=0.01em]
    \item We provide the first comprehensive investigation of standard training strategies of CBMs for interpreting PLMs and benchmark CBE-PLMs.
    \item We propose C$^3$M, which leverages LLMs and MixUp to help PLMs learn from human-annotated and machine-generated concepts. C$3$M liberates CBMs from predefined concepts and enhances the interpretability-utility tradeoff.
    \item We demonstrate the effectiveness and robustness of test-time concept intervention for the learned CBE-PLMs for common text classification tasks.
\end{itemize}


\section{{Related Work}}
\subsection{Interpreting Pretrained Language Models}
PLMs such as Word2Vec~\cite{mikolov2013efficient}, BERT~\cite{devlin2018bert}, and the more recent GPT series~\cite{radford2019language,brown2020language,openai2023gpt4} have demonstrated impressive performance in various NLP tasks. However, their opaque nature poses a challenge in comprehending how PLMs work internally~\cite{diao2022black}. In order to improve the interpretability and transparency of PLMs, researchers have explored different approaches, such as visualizing attention weights~\cite{galassi2020attention}, probing feature representations~\cite{mishra2017local,lundberg2017unified,bills2023language}, and using counterfactuals~\cite{wu2021polyjuice,ross2021explaining}, among others, to provide explanations at the local token-level, instance-level, or neuron-level. However, these methods often lack faithfulness and intuitiveness, and are of poor readability, which undermines their trustworthiness~\cite{madsen2022post}.

Recently, researchers have turned to global concept-level explanations that are naturally understandable to humans. Although this level of interpretability has been less explored in NLP compared to computer vision ~\cite{goyal2019explaining,kim2018interpretability,mu2020compositional}, it has gained attention. For instance, a study~\cite{vig2020investigating} investigates gender classification bias by examining the association of occupation words such as `nurse' with gender. In addition, the CBMs~ \cite{koh2020concept,zarlenga2022concept} have emerged as novel frameworks for achieving concept-level interpretability in lightweight image classification systems.
CBMs typically involve a layer preceding the final fully connected classifier, where each neuron corresponds to a concept that can be interpreted by humans.
CBMs also show advantages in improving accuracy through human intervention during testing. Yet, the application of CBMs to larger-scale PLMs interpretation is under-explored. Implementing CBMs necessitates human involvement in defining the concept set and annotating the concept labels. Such requirements are challenging for natural language as humans may need to read through the entire text to understand the context and label one concept~\cite{nemeth2020machine}.

\subsection{Learning from Noisy Labels}
Addressing inaccurately labeled or misclassified data in real-world scenarios is the goal of learning from noisy labels, with techniques including noise transition matrix estimation~\cite{liu2022identifiability}, robust risk minimization~\cite{englesson2021generalized}, and more. 
Recently, the resilience of semi-supervised learning methods like MixMatch~\cite{berthelot2019mixmatch} and FixMatch~\cite{sohn2020fixmatch} to label noise has been discovered by using pseudo-labels for unlabeled data. 
Inspired by them, we porpose to utilize an LLM (ChatGPT) as a fixed-label guesser, generating noisy intermediate concept labels to potentially predict task labels.

Notably, CBMs specialize in the interpretation and interactability of deep models for general classification tasks. While \textit{Multi-Aspect Sentiment Analysis}~\cite{zhang2022survey} (MASA) shares similar goals when using aspects as concepts, it differs as concepts are not confined to fine-grained aspectual features and can be abstract ideas or broader notions throughout entire contexts. Aspect labels in MASA, primarily used for prediction accuracy, are not always mandatory. 
To summarize, this study pioneers the comprehensive exploration of utilizing concepts for interpreting large-scale PLMs, and provids a robust framework for harnessing the noisy signals from LLMs to achieve interpretable outcomes from lighter-weight PLMs, which can be easily understood by users.



\begin{figure*}[htbp]
  \centering\scalebox{0.93}{
  \includegraphics[width=\linewidth]{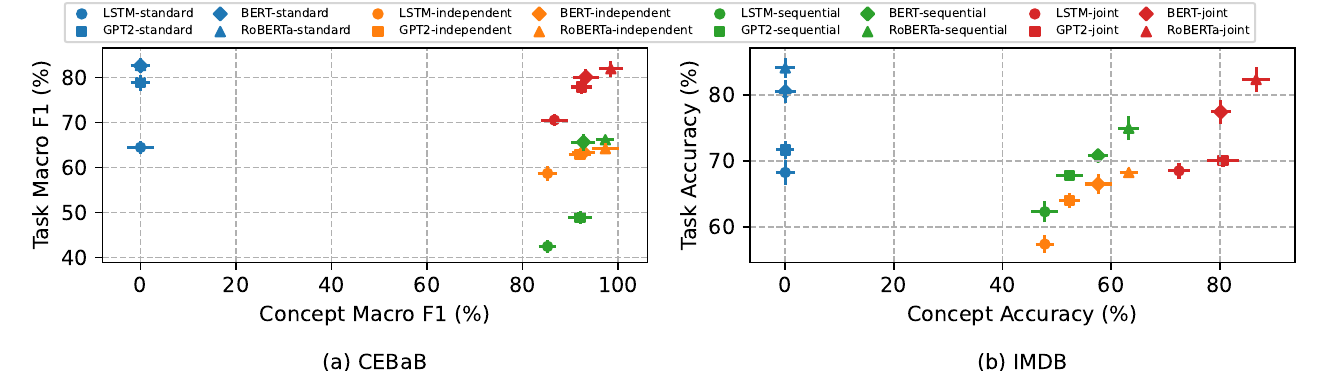}}
  \caption{Illustration of the interpretability-accuracy trade-off using various backbones. The top-right indicates a more favorable trade-off, i.e., a better interpretability-utility Pareto front. We also show the confidence intervals for both dimensions.}
  \label{fig:tradeoff}
  \vspace{-0.5cm}
\end{figure*}

\section{Enable Concept Bottlenecks for Pretrained Language Models}
\subsection{Problem Setup}\label{sec:setup}
We focus on interpreting the predictions of fine-tuned PLMs for text classification tasks. Given data $\mathcal{D} = \{(x^{(i)}, y^{(i)}, c^{(i)})_{i=1}^n\}$, where $x \in \mathbb{R}^d$ is the original text input, $y \in \mathbb{R}$ is the target label to predict, and $c \in \mathbb{R}^k$ is a vector of $k$ concepts from the concept set $\mathcal{C}$ with $|\mathcal{C}| = k$, we consider a PLM $f_\theta$ parameterized by $\theta$ that encodes an input text $x \in \mathbb{R}^d$ into its latent representation $z \in \mathbb{R}^e$. Vanilla fine-tuning strategy, concretely defined in Appendix~\ref{app:def}, can be abstracted as $x \rightarrow z \rightarrow y$.

\textbf{Concept-Bottleneck-Enabled Pretrained Language Models.} The original concept bottlenecks in CBMs~\citep{koh2020concept} come from resizing one of the layers in the CNN encoder to match the number of concepts. However, since PLM encoders typically provide text representations with much higher dimensions than the number of concepts, directly reducing the neurons in the layer would significantly impact the quality of learned text representation. To address this issue, we instead add a linear layer with the sigmoid activation, denoted as $p_\psi$, that projects the learned latent representation $z \in \mathbb{R}^e$ into the concept space $c \in \mathbb{R}^k$. This process can be represented as $x \rightarrow z \rightarrow c \rightarrow y$. Note that, unlike the previous works for image classification, each concept here does not need to be binary (i.e., present or not). We allow multi-class concepts, e.g., the concept ``Food'' in a restaurant review can be positive, negative, or unknown. We refer to the PLM and the projector $(f_\theta, p_\psi)$ together as the \textit{concept encoder} and the complete model~$(f_\theta, p_\psi, g_\phi)$ as \textit{Concept-Bottleneck-Enabled Pretrained Language Models} (CBE-PLMs).

During training, CBE-PLMs seek to achieve two goals: (1) align concept prediction $\hat{c}=p_\psi(f_\theta(x)))$ to $x$’s ground-truth concept labels $c$ and (2) align label prediction $\hat{y}=g_\phi(p_\psi(f_\theta(x)))$ to ground-truth task labels $y$. We accordingly adapt the three conventional strategies, \textit{independent} training, \textit{sequential} training, and \textit{joint} training, proposed in~\citep{koh2020concept} to learn the CBE-PLM. Their detailed formulations are given in Appendix~\ref{app:def}.


\subsection{Benchmarking CBE-PLMs}
We propose to benchmark the performance of the vanilla fine-tuning and the three training strategies for CBE-PLMs using two text classification datasets: \texttt{CEBaB}~\citep{abraham2022cebab} and \texttt{IMDB}~\citep{maas2011learning}. Both datasets contain human-labeled concepts.
We consider four typical PLMs following~\citet{abraham2022cebab}.
Descriptions of the PLM backbones, datasets, and concept labels are detailed in Section~\ref{sec:dataset}, Section~\ref{sec:backbone}, and Appendix~\ref{app:trans_con}. We consider the target task scores and concept prediction scores as the evaluation metrics for utility and interpretability, respectively.

\textbf{CBM for CBE-PLMs}.
In this experiment, we aim to identify the optimal training strategy for CBE-PLMs. The results depicted in Figure~\ref{fig:tradeoff} confirm that standard-PLMs typically yield the highest task scores, demonstrating that the implementation of a concept bottleneck can indeed impact target task performance negatively. However, without considering the concept labels, standard-PLMs lack interpretability. In contrast, CBE-PLMs trained jointly exhibit higher task scores and superior concept prediction scores compared to their counterparts. This divergence from CBMs in the image domain, where all three strategies display similar performance~\citep{koh2020concept}, is notable. We attribute this to PLMs' extensive pretraining on numerous human-generated corpora and larger parameter numbers than the studied vision encoders such as ResNets~\citep{he2016deep}. Unlike independent or sequential training where the PLM encoder is fixed after training on the concept labels, joint training allows PLMs to ultize their capacity to learn concepts and target labels jointly, making the learned concept activations from the bottleneck layer better aligned with the task labels. Given this advantage of joint training, we adopt it as the default strategy for training CBE-PLMs in the subsequent sections.

While initial findings from applying vanilla CBM~\citep{koh2020concept} for interpreting PLMs appear encouraging, they require human-annotated concepts during training. This proves to be impractical in real-world situations due to the vast number of potential concepts and the time-intensive annotation process~\citep{nemeth2020machine}. Often, only a limited number of texts come with manually labeled concepts. Moreover, as humans continuously acquire new concepts, it is desirable for the training framework to discover and incorporate new concepts automatically. Thus, we aim to design a general framework for training CBE-PLMs.

\section{C$^{3}$M: A General Framework for Learning CBE-PLMs}
 
\begin{figure}[htbp]
  \centering\scalebox{0.7}{
  \includegraphics[width=\linewidth]{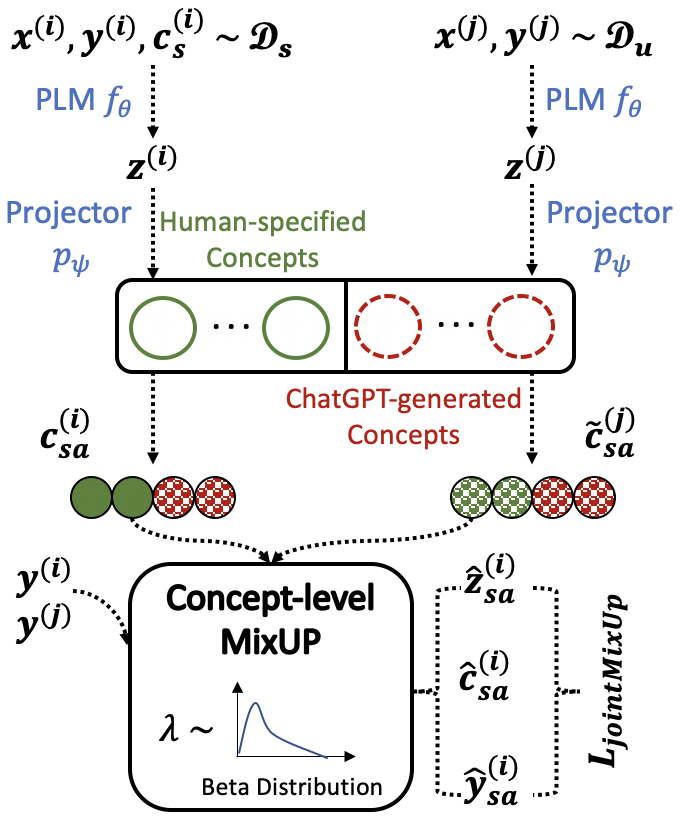}}
  \caption{Illustration of the proposed framework C$^3$M.}
  \label{fig:c3m}
  \vspace{-0.5cm}
\end{figure}

We define the following data potions according to the real-world scenarios. We refer to a dataset with human-annotated concepts as the \textit{source concept dataset}, denoted as $\mathcal{D}_s = \{(x^{(i)}, y^{(i)}, c_s^{(i)})_{i=1}^{n_s}\}$, where $n_s$ denotes the size and $c_s \in \mathbb{R}^{k_s}$ is a vector of $k_s$ concepts from the pre-defined source concept set $\mathcal{C}_s$. We also consider another dataset without concept labels, referred to as the \textit{unlabeled concept dataset}, denoted as $\mathcal{D}_u = \{(x^{(i)}, y^{(i)})_{i=1}^{n_u}\}$. The complete dataset is then the combination of these two datasets: $\mathcal{D} = \{\mathcal{D}_s, \mathcal{D}_u\}$. $n_s$ and $k_s$ are typically small, limiting the effectiveness of CBE-PLMs. Specifically, small $n_s$ leads to sparse concept labels in $\mathcal{D}$, and vanilla CBM cannot be trained on datasets with unlabeled concepts $\mathcal{D}_u$. Additionally, small $k_s$ indicates that we may not have sufficient information for model prediction. To address these limitations, we propose \textit{\underline{C}hatGPT-guided \underline{C}oncept augmentation with \underline{C}oncept-level \underline{M}ixup} (C$^3$M), a novel framework for training CBE-PLMs effectively. As illustrated in Figure~\ref{fig:c3m}, at the high level, we augment the concept set $\mathcal{C}_s$ and annotate pseudo concept labels for the unlabeled concept dataset using ChatGPT. Since these pseudo labels are noisy, we propose a novel concept-level MixUp to train the CBE-PLMs effectively on the augmented dataset with noisy concept labels. 

\subsection{ChatGPT-guided Concept Augmentation}\label{sec:aug_c}
In this section, we detail how to leverage ChatGPT (GPT4) to automatically (1) augment the concept set, and (2)  annotate missing concept labels.
\subsubsection{Concept Set Augmentation}\label{sec:csa}
The goal of concept set augmentation is to automatically generate high-quality concepts using human-specified concepts $\mathcal{C}_s$ as references. These generated concepts should be semantic meaningful and useful for target task prediction. Inspired by LF-CBM~\citep{oikarinenlabel}, we query ChatGPT with appropriate prompts to generate additional concepts. Our prompts are designed using ``in-context learning''~\citep{brown2020language,min2022rethinking,xieexplanation}, and include examples from human annotations. Below is an example of a ChatGPT prompt designed for a sentiment classification task using the IMDB dataset \citep{maas2011learning}:

\begin{mdframed}[backgroundcolor=gray!20]
\small
Besides $\{Acting$, $Storyline$, $Emotional$ $Arousal$, $Cinematography\}$, what are the additional important features to judge if a $\{movie\}$ is good or not?
\end{mdframed}

Parentheses represent fields that can be customized for different tasks. The concepts $Acting$, $Storyline$, $Emotional$ $Arousal$, and $Cinematography$ are from the source concept set $\mathcal{C}_s$ with labels manually annotated following procedures in Appendix~\ref{app:anno}. Different from LF-CBM which generates concepts merely relying on GPT3~\citep{brown2020language}, we further include a small set of human-specified concepts in the prompt to improve the quality of generated concepts. This additional information can help effectively filter out undesired output without additional operations (e.g., deletions). Rarely seen concepts are discarded using a predefined threshold and the remaining generated concepts are referred to as \textit{augmented concepts set} $\mathcal{C}_a$ with size $k_a$. Results are given in Table~\ref{tab:trans_con} in Appendix~\ref{app:trans_con}.
\subsubsection{Noisy Concept Label Annotation}
The next step is to automatically annotate unlabeled concepts using \textit{noisy} labels. We again leverage the power of ChatGPT which has been shown to encapsulate significant amounts of human common sense knowledge~\citep{bommasani2022opportunities,openai2023gpt4,singh2023mind} and show strong performance for some text annotation tasks \cite{gilardi2023chatgpt}. As we will also show here, LLMs are surprisingly proficient at identifying language concepts when suitably prompted. Using the same example of movie reviews, the prompt for this step is designed as follows:
\begin{mdframed}[backgroundcolor=gray!20]
\small
a. According to the review "$\{text_1\}$", the "$\{concept_1\}$" of the movie is "positive".\\
b. According to the review "$\{text_2\}$", the "$\{concept_2\}$" of the movie is "negative".\\
c. According to the review "$\{text_3\}$", the "$\{concept_3\}$" of the movie is "unknown".\\
d. According to the review "$\{text_i\}$", how is the "$\{concept_i\}$" of the movie? Please answer with one option in "positive, negative, or unknown".
\end{mdframed}
Following a similar ``in-context learning'' strategy described in Section~\ref{sec:csa}, Prompts a-c are three human-annotated examples randomly selected to represent positive, negative, and unknown concept labels, respectively. Prompt d is the query instance. The goal is to obtain noisy labels for any given $\{text_i\}$ and $\{concept_i\}$. There are three types of noisy concept annotations:
\begin{enumerate}[leftmargin=*, itemsep=0.01em]
    \item Noisy labels for human-specified concepts in $\mathcal{D}_s$. The resulting dataset $\Tilde{\mathcal{D}}_{s}$ is used to validate the quality of labels generated by ChatGPT only (See Table~\ref{tab:stats_concept_s} in Appendix~\ref{app:stat_hac}). 
    \item Noisy labels for ChatGPT-generated concepts in $\mathcal{D}_s$. The augmented concept set is denoted as $c_{sa}=(c_s || c_a)\in \mathbb{R}^{k_s+k_a}$, where $||$ refers to the concatenation operator and $c_a \in \mathbb{R}^{k_a}$ stands for the generated concepts. For example, we identify new important concepts such as \textit{Soundtrack} using ChatGPT for the IMDB movie reviews.
    \item Noisy labels for both human-specified and ChatGPT-generated concepts in unlabeled concept datasets $\mathcal{D}_u$. The augmented concept set is denoted as $\Tilde{c}_{sa}=(\Tilde{c_s} || \Tilde{c_a})\in \mathbb{R}^{k_s+k_a}$ and $\Tilde{c_s} \in \mathbb{R}^{k_s}, \Tilde{c_a} \in \mathbb{R}^{k_a}$ stand for the generated concept labels for human-specified and ChatGPT-generated concepts, respectively. 
\end{enumerate}
In summary, we transform the original dataset with sparse concept labels into an augmented dataset with new concepts and noisy labels: $\mathcal{D} = \{\mathcal{D}_s, \mathcal{D}_u\} \rightarrow \Tilde{\mathcal{D}} = \{\Tilde{\mathcal{D}}_{sa}, \Tilde{\mathcal{D}}_u\}$. 
Examples of these two types of qeuries are illustrated in Appendix~\ref{app:gpt}.

\vspace{-0.1cm}
\subsection{Learning from Noisy Concept Labels}
While directly training CBE-PLMs on the transformed dataset $\Tilde{\mathcal{D}}$ is straightforward, this method's drawback is its equal treatment of human annotations and ChatGPT-generated noisy labels, potentially leading to prediction and interpretation inaccuracies. To improve interpretability and accuracy, we introduce a novel \textit{Concept-level MixUp} (CM) approach. It advocates for a convex behavior of PLMs between human-annotated and ChatGPT-generated concepts, thereby enhancing its robustness against noisy concept labels.


\subsubsection{Concept-level MixUp}
To better utilize the noisy concept labels, CM first linearly interpolates the texts and concept labels between human-annotated concepts ($\Tilde{\mathcal{D}_{sa}}$) and ChatGPT-generated concepts ($\Tilde{\mathcal{D}_u}$). Specifically, we interpolate any two text-concept-label ternaries $(x^{(i)}, c^{(i)}, y^{(i)})$, $(x^{(j)}, c^{(j)},  y^{(j)})$ for both their latent representation ($z^{(i)}, z^{(j)}$), concepts ($c^{(i)}, c^{(j)}$), and the task labels ($y^{(i)}, y^{(j)}$) using the MixUp ($\cdot$) defined as follows:
\begin{equation}\label{eq:mixup}
\begin{aligned}
    \lambda \sim\text{Beta}(\alpha,\alpha) &;\ \ 
    \hat{\lambda} = \max(\lambda, 1-\lambda); \\
    z^{(i)} = f_{\theta} (x^{i}) &;\ \  
    z^{(j)} = f_{\theta} (x^{j}); \\ 
    {\hat{z}}^{(i,j)} = \hat{\lambda} z^{(i)} &+ (1-\hat{\lambda}) z^{(j)}; \\ 
    {\hat{c}}^{(i,j)} = \hat{\lambda} c^{(i)} &+ (1-\hat{\lambda}) c^{(j)};\\ 
    {\hat{y}}^{(i,j)} = \hat{\lambda} y^{(i)} &+ (1-\hat{\lambda}) y^{(j)},
\end{aligned}
\end{equation}
where $\alpha$ is a hyperparameter for the Beta distribution. Notably, $\hat{\lambda} \geq 0.5$ preserves the order of human-annotated concepts and ChatGPT-generated concepts for computing individual loss components in Eq.~\eqref{eq:loss} appropriately. Then, we combine and shuffle human-annotated and ChatGPT-annotated data {in the transformed dataset $\Tilde{\mathcal{D}} = \{\Tilde{\mathcal{D}}_{sa}, \Tilde{\mathcal{D}}_u\}$:
\begin{equation}
    \mathcal{W} = \text{Shuffle}(\Tilde{D}) = \text{Shuffle}(\Tilde{\mathcal{D}}_{sa}||\Tilde{\mathcal{D}}_u),
\end{equation}
where $||$ indicates the concatenation of two potions of datasets. Next, we perform MixUp ($\cdot$) for the $i$th instance as follows:
\begin{equation}
    \begin{aligned}
    ({\hat{z}_{sa}}^{(i)}, {\hat{c}_{sa}}^{(i)}, {\hat{y}_{sa}}^{(i)})&=\text{MixUp}(\Tilde{\mathcal{D}}_{sa}^{(i)}, \mathcal{W}^{(i)}), \\
     ({\hat{z}_{u}}^{(i)}, {\hat{c}_{u}}^{(i)}, {\hat{y}_{u}}^{(i)}) &= \text{MixUp}(\Tilde{\mathcal{D}}_{u}^{(i)}, \mathcal{W}^{(i)}).
    \end{aligned}
\end{equation} 
Through these steps, we can generate a "mixed version" for each instance in $\Tilde{D}_{sa}$ and $\Tilde{D}_{u}$, while preserving a larger portion of the original instance.

\subsubsection{Loss Function}
The loss function $L_{jointMixUp}$ for training CBE-PLMs with the MixUped dataset is defined below:
\begin{equation}\label{eq:loss}
    \begin{aligned}
        L_{sa} &=L_{joint} ({\hat{z}_{sa}}^{(i)},\hat{c}_{sa}^{(i)}, \hat{y}_{sa}^{(i)}); \\ 
        L_{u} &= L_{joint} ({\hat{z}_{u}}^{(i)},\hat{c}_{u}^{(i)}, \hat{y}_{u}^{(i)}); \\ 
        L_{jointMixUp} &= L_{sa} + \tau L_{u} ,
    \end{aligned}
\end{equation}
where $\tau$ is a hyperparameter and $L_{joint}$ is the joint training loss used in vanilla CBM formulated in Appendix~\ref{app:def}. In this way, We backpropagate gradients of the mixed noisy concept labels and gold concept labels to update the parameters in CBE-PLMs. 

\section{Experiments}
\subsection{Datasets}\label{sec:dataset}
In this section, we give detailed descriptions of the experimented datasets. Each of the datasets has two components: source concept dataset and unlabeled concept dataset ($\mathcal{D} = \{\mathcal{D}_s, \mathcal{D}_u\}$). Existing datasets with human-annotated concept labels are very limited. One source concept dataset is \texttt{CEBaB}~\citep{abraham2022cebab,wu2022causal}, a common sentiment classification dataset for restaurant reviews. Its corresponding $\mathcal{D}_u$ is the restaurant reviews from the Yelp Dataset\footnote{\url{https://www.kaggle.com/datasets/omkarsabnis/yelp-reviews-dataset}}. We also curate another dataset for movie reviews. Specifically, we randomly sample two portions of reviews from the IMDB datasets~\citep{maas2011learning} to represent $\mathcal{D}_s$ and $ \mathcal{D}_u$, respectively. Following a previous NLP work~\citep{cai2021aspect}, we manually annotate the concept labels for $\mathcal{D}_s$ in the movie reviews. More annotation details are included in Appendix~\ref{app:anno}. For convenience, we still refer to these two new datasets as \texttt{CEBaB} and \texttt{IMDB}. Each concept contains three values, i.e., Negative, Positive, and Unknown. As described in Section~\ref{sec:aug_c}, each dataset $\mathcal{D}$ is then transformed into $\Tilde{\mathcal{D}} = \{\Tilde{\mathcal{D}_{sa}}, \Tilde{\mathcal{D}_u\}}$. The basic statistics of the transformed datasets and their human-annotated concepts are given in Table~\ref{tab:stats_data} in Appendix~\ref{app:split}  and  Table~\ref{tab:stats_concept_s} in Appendix~\ref{app:stat_hac}, respectively. Note that the last column in Table~\ref{tab:stats_concept_s} indicates the accuracy of ChatGPT-labeled concepts in $\mathcal{D}_s$, as described in Section~\ref{sec:aug_c}. Table~\ref{tab:trans_con} in Appendix~\ref{app:trans_con} provides statistics about augmented concepts. Both the human-annotated and ChatGPT-generated data, alone with the framework implementation are released~\footnote{\url{https://github.com/Zhen-Tan-dmml/CBM\_NLP.git}}.


\subsection{PLM Backbones}\label{sec:backbone}
We experiment with the same PLM backbones as in the CEBaB paper~\cite{abraham2022cebab}: GPT2~\citep{radford2019language}, BERT~\citep{devlin2018bert}, RoBERTa~\citep{liu2019roberta}, and BiLSTM~\citep{hochreiter1997long} with CBOW~\citep{mikolov2013efficient}. For better performance, we obtain the representations of the input texts by pooling the embedding of all tokens. Reported scores are the averages of six independent runs, each taking 5 to 40 minutes. More implementation details and parameter values are included in Appendix~\ref{app:implement} and Table~\ref{tab:symbols} in Appendix~\ref{app:para}.

%
\begin{table*}[htbp]
\caption{{Comparisons of task accuracy and interpretability using \texttt{CEBaB} and \texttt{IMDB} datasets.} Metrics for both task and concept labels are written as \textbf{Accuracy}/\textbf{Macro F1}. Scores are reported in $\%$. Scores in \textbf{bold} indicate that the CBE-PLM under the current setting outperforms its standard PLM counterpart. CM denotes Concept-level MixUp.\label{tab:compare}}
\vspace{-0.3cm}
\begin{center}
\scalebox{0.67}{
\begin{tabular}{@{}cc|cccc|cccc@{}}
\toprule
\multicolumn{2}{c|}{\textbf{Dataset}}                      & \multicolumn{4}{c|}{\texttt{CEBaB}}                                              & \multicolumn{4}{c}{\texttt{IMDB}}                                            \\ \midrule
\multicolumn{2}{c|}{\multirow{2}{*}{\textbf{Model}}}       & \multicolumn{2}{c}{$\mathcal{D}$}              & \multicolumn{2}{c|}{$\Tilde{\mathcal{D}}$}            & \multicolumn{2}{c}{$\mathcal{D}$}          & \multicolumn{2}{c}{$\Tilde{\mathcal{D}}$}             \\ \cmidrule(l){3-10} 
\multicolumn{2}{c|}{}                                      & Task                 & Concept     & Task                 & Concept     & Task               & Concept   & Task                 & Concept     \\ \midrule
\multicolumn{1}{c|}{\multirow{4}{*}{PLMs}}       & LSTM    & 40.57/60.67          & -           & 43.34/64.47          & -           & 68.25/53.37        & -         & 90.5/90.46           & -           \\
\multicolumn{1}{c|}{}                            & GPT2    & 66.69/77.25          & -           & 67.26/78.81          & -           & 71.67/67.53        & -         & 97.64/97.55          & -           \\
\multicolumn{1}{c|}{}                            & BERT    & 68.75/78.71          & -           & 71.81/82.58          & -           & 80.5/78.4          & -         & 98.89/98.68          & -           \\
\multicolumn{1}{c|}{}                            & RoBERTa & 71.36/80.17          & -           & 73.12/82.64          & -           & 84.1/82.5          & -         & 99.13/99.12          & -           \\ \midrule
\multicolumn{1}{c|}{\multirow{4}{*}{CBE-PLMs}}   & LSTM    & \textbf{56.47/67.82} & 86.46/85.24 & \textbf{54.54/65.84} & 83.46/84.74 & \textbf{68.5/55.4} & 72.5/77.5 & \textbf{93.02/91.53} & 76.92/75.41 \\
\multicolumn{1}{c|}{}                            & GPT2    & 64.04/77.75          & 92.14/92.05 & 63.57/74.71          & 90.17/90.13 & 70.05/69.53        & 80.6/82.5 & 96.85/96.81          & 86.14/88.06 \\
\multicolumn{1}{c|}{}                            & BERT    & 67.27/79.24          & 93.65/92.75 & 68.23/78.13          & 89.64/90.45 & 77.42/74.57        & 80.2/83.7 & 97.62/97.58          & 92.57/92.05 \\
\multicolumn{1}{c|}{}                            & RoBERTa & 70.98/79.89          & 96.12/95.34 & 69.85/79.29          & 91.45/92.23 & 82.33/80.13        & 86.7/85.3 & 98.45/98.12          & 93.99/94.28 \\ \midrule
\multicolumn{1}{c|}{\multirow{4}{*}{CBE-PLMs-CM}} & LSTM    & -                    & -           & \textbf{59.67/70.53} & 88.75/86.67 & -                  & -         & \textbf{94.35/92.32} & 83.83/84.52 \\
\multicolumn{1}{c|}{}                            & GPT2    & -                    & -           & 65.54/77.87          & 93.58/92.32 & -                  & -         & \textbf{97.89/97.88} & 89.64/88.25 \\
\multicolumn{1}{c|}{}                            & BERT    & -                    & -           & 70.58/80.07          & 94.43/93.26 & -                  & -         & 98.18/98.06          & 94.87/94.32 \\
\multicolumn{1}{c|}{}                            & RoBERTa & -                    & -           & 72.88/81.91          & 96.3/98.5   & -                  & -         & \textbf{99.69/99.66} & 96.35/96.36 \\ \bottomrule
\end{tabular}}
\end{center}
\vspace{-0.5cm}
\end{table*}

\subsection{Task Accuracy vs Interpretability}
Table~\ref{tab:compare} presents the results for the two original datasets ($\mathcal{D}$) and their transformed versions ($\Tilde{\mathcal{D}}$). We have the following observations:

\textbf{{CBE-PLMs offer interpretability and competitive task prediction performance.}} Compared to standard PLMs (trained solely with task labels), CBE-PLMs provide concept-level interpretability with only a minor decrease in task prediction. Interestingly, a smaller PLM, i.e., LSTM with CBOW embeddings, achieves improved task accuracy when learning from concept labels. This suggests that the accuracy-interpretability tradeoff in concept learning is not necessary, as opposed to the prevailing view. Concepts can help guide PLMs trained on smaller corpora with fewer parameters towards better prediction performance.

\textbf{Noisy concept labels can facilitate the training of CBE-PLMs on small datasets.} The extremely limited size of the \texttt{IMDB} source concept dataset (deliberately set to 100) yields unsurprisingly low test scores. 
Transforming $\mathcal{D}$ into $\Tilde{\mathcal{D}}$ using ChatGPT for noisy labeled concept instances leads to significant improvements in both concept and task predictions for CBE-PLMs-CM.

\textbf{Uncritical learning from noisy concept labels can impair performance.} Results for \texttt{CEBaB} in Table~\ref{tab:compare} demonstrate that, learning from the transformed dataset $\Tilde{\mathcal{D}}$ directly leads to inferior performance for CBE-PLMs. Unlike \texttt{IMDB}, the source concept dataset in \texttt{CEBaB} contains sufficient training instances, therefore, enforcing CBE-PLMs to learn from noisy concept labels will undesirably mislead the model, exacerbating both the concept and task prediction performance.

\textbf{CBE-PLMs-CM trained via the proposed C$^3$M framework consistently deliver superior interpretability-utility trade-offs.} By encouraging the CBE-PLMs to linearly interpolate between examples with gold-labeled concepts and those with ChatGPT-generated concepts, the model is able to extract useful semantic knowledge meanwhile becoming robust to noisy concept labels. The result is promising: We achieve the best concept-level prediction (interpretability measure) without sacrificing the task prediction performance, and in some cases, CBE-PLMs trained through C$^3$M can even outperform their standard PLM counterparts. 

\subsection{Explainable Predictions}
A unique advantage of CBMs is that its decision rules can be interpreted as a linear combination of comprehensible variables~\citep{koh2020concept}. Inheriting this strength, our proposed CBE-PLMs can deliver intuitive concept-level explanations for predictions by assessing the activations of each concept. We measure concept contribution using the product of activation and the corresponding weight in the linear label predictor $g_\phi$~\cite{oikarinenlabel}.
Concepts with negative activation are designated as ``Neg Concept''. We highlight the concepts contributing the most in our visualizations. Visualization results are demonstrated in Figure~\ref{fig:logit_0} for a toy example, while real-world \texttt{CEBaB} and \texttt{IMDB} case studies can be found in Appendix~\ref{app:logits}. These visualizations provide new intriguing insights into real-world applications. 
For instance, negative concepts (e.g., Service) contribute more to the final prediction of positive sentiment in Figure~\ref{fig:logit_0}, making the predicted sentiment second highest ($Y=4$) rather than the highest ($Y=5$). 
Moreover, interpretability results such as Figure~\ref{fig:logit_1} in Appendix~\ref{app:logits} imply that concepts such as ``Food'' and ``Ambiance'' weigh more heavily in customers' restaurant evaluations compared to ``Noise'' and ``Menu Variety''.


\begin{figure}[htbp]
\vspace{-0.2cm}
  \scalebox{0.9}{
  \includegraphics[width=\linewidth]{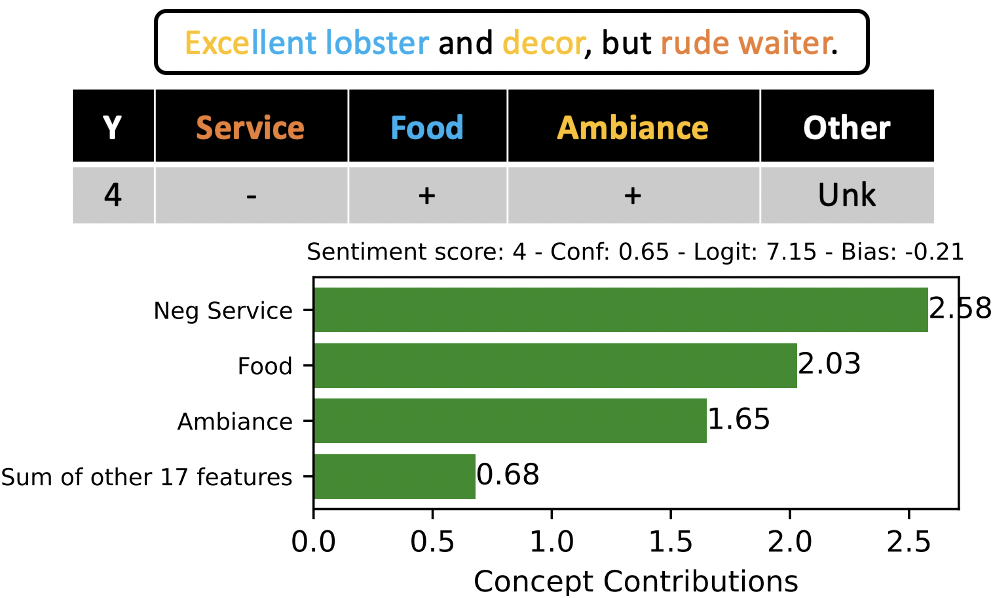}}
  \caption{{Illustration of the explainable prediction for a toy example in restaurant review sentiment analysis.}}
  \label{fig:logit_0}
  \vspace{-0.4cm}
\end{figure}



\subsection{Test-time Intervention}

\begin{figure}[htbp]
\vspace{-0.15cm}
		\centering
\captionsetup[sub]{skip=-0.5pt}
\subcaptionbox{BERT}
{\includegraphics[width=0.235\textwidth]{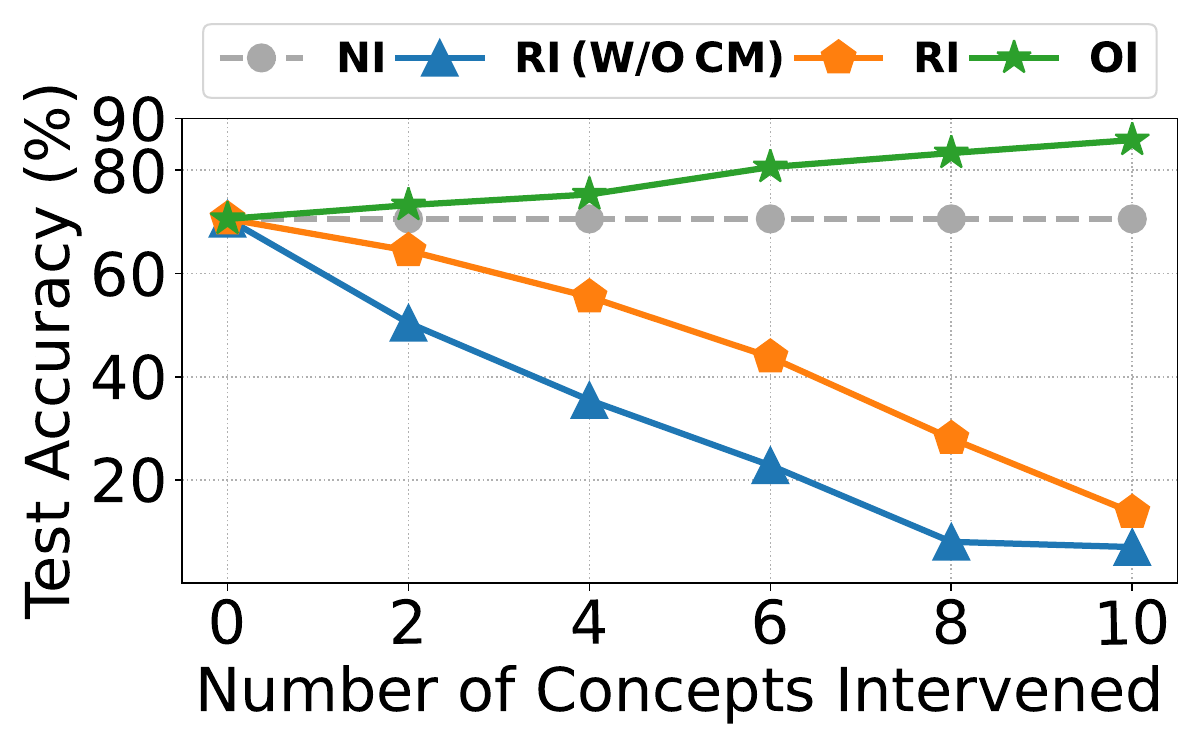}}
\subcaptionbox{{GPT2}}
{\includegraphics[width=0.235\textwidth]{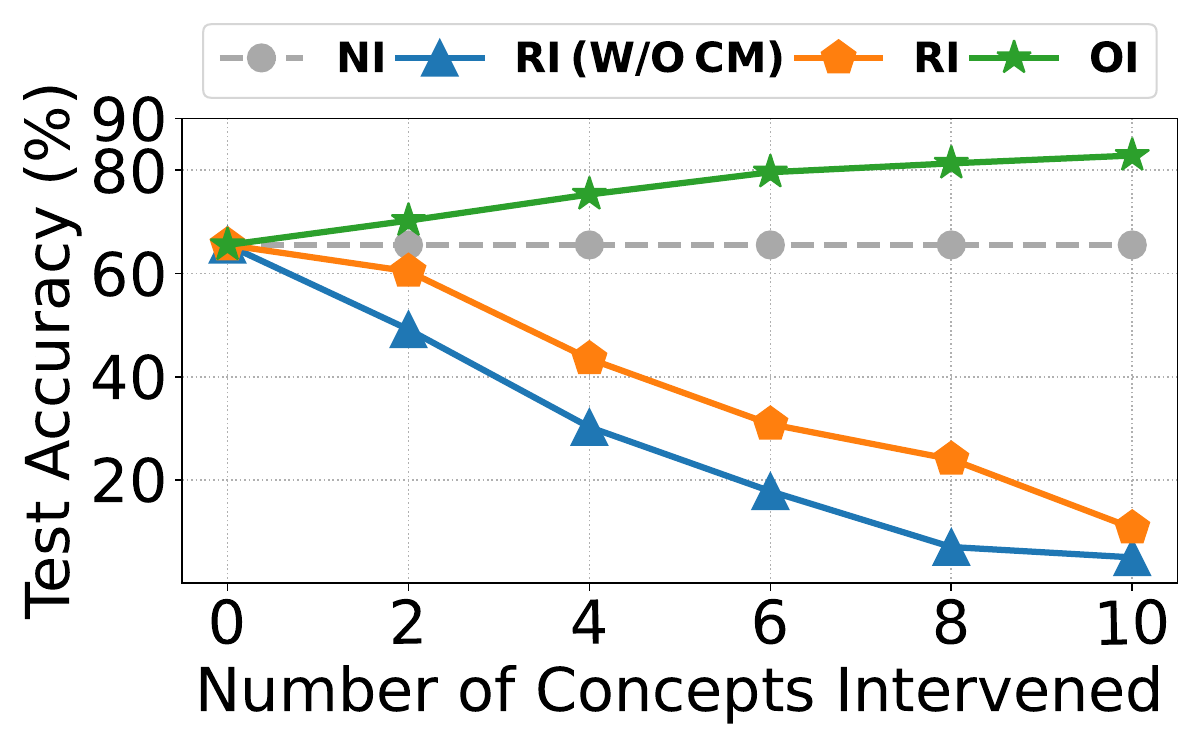}}
\vspace{-0.1in}
		\caption{The results of Test-time Intervention. "NI" denotes "no intervention", "RI (W/O CM)" denotes "random intervention on CBE-PLMs without the concept-level MixUp", "RI" denotes "random intervention on CBE-PLMs", and "OI" denotes "oracle intervention".}
		\label{fig:intervene}
\vspace{-0.11in}
	\end{figure}
Another strength of CBE-PLMs is that they allow test-time concept intervention (inherited from CBMs), facilitating deeper, user-friendly interactions. {To assess this strength, we follow \citet{koh2020concept} to intervene in the predicted concepts and investigate the impact of such interventions on test-time prediction accuracy}. 
Concept mispredictions arise from ChatGPT's incorrect labels or inaccurate concept activation. 
Recall that the input of the task label predictor is the predicted concept activations $\hat{a} = p_\phi(f_\theta(x))$ rather than the predicted ternary concepts $\hat{c}$. In a concept-level intervention $I$, the activation $\hat{a}_j$ of the $j$th concept with a target concept $c_j$ is set to the 5th, 95th, or 50th percentile of $\hat{a}_j$ over the training distribution for Negative, Positive, or Unknown $c_j$ respectively. Multiple concepts can be intervened upon by replacing all related predicted concept activations and updating the prediction. Experiments were conducted on the transformed version {$\Tilde{\mathcal{D}}$} of the \texttt{CEBaB} dataset. Figure~\ref{fig:intervene} exhibits results for CBE-PLMs using BERT and GPT2 as the PLM backbones (with similar observations for LSTM and RoBERTa). {A case study is further illustrated in Appendix~\ref{app:internvene}}. The results reveal that task accuracy improves substantially when more concepts are corrected by the oracle. Additionally, while the performance of CBE-PLMs declines as more concepts are intervened upon incorrectly (randomly), the proposed concept-level MixUp effectively mitigates this impact. Notably, the decline in performance is marginal when only two concepts are erroneously intervened upon. These findings underscore the pronounced advantages of test-time intervention for CBE-PLMs trained through C$^3$M. First, domain experts can interact with the model to rectify any inaccurately predicted concept values. Second, in reality, even experts might inadvertently implement incorrect interventions. Yet, despite this susceptibility, our proposed concept-level MixUp strategy effectively curbs performance degradation, particularly when inaccuracies affect only a small subset of the intervention. This attests to the robustness of the proposed framework.


\section{Conclusion} 
Our analysis began with an exhaustive examination of three training strategies, identifying joint training as the most efficacious. Further, we proposed the C$^3$M framework, designed to streamline the training process of CBE-PLMs in the presence of incomplete concept labels. Moreover, we showcased the interpretability of our models in their decision-making process and elucidated how this comprehensibility can be harnessed to boost test accuracy via concept intervention. 

\textbf{Outlook: }Our research lays the groundwork for future studies focused on enhancing the transparency and robustness of PLMs. We foresee that CBE-PLMs could potentially show more resilience to data biases compared to standard PLMs, which have been known to display biased performance due to spurious correlations between sensitive attributes (e.g., gender) and task labels~\cite{wang2021robustness,udomcharoenchaikit2022mitigating}. For instance, a biased PLM might wrongly infer patterns like female users writing more extreme reviews while male users tend towards moderate ones. CBE-PLMs, by focusing on concept labels and relying solely on these concepts for classifications, might reduce such biases. If the concepts are not associated with sensitive attributes and their relationship with task labels is consistent, CBE-PLMs could offer enhanced fairness.

\section*{Limitations}
While our approach presents a significant step towards more interpretable pretrained language models, several limitations warrant further exploration.
First, our approach relies heavily on the accuracy of the predefined concepts. Despite the promising results, this dependency raises the issue of potential bias present in the concept selection process (for both human-specified and ChatGPT-generated concepts). If a concept is not well-defined or if important concepts are missing, this could lead to incomplete or skewed interpretations.
Second, the methodology proposed in this paper has not been experimented on very large language models, such as Bloom~\cite{scao2022bloom}. The core idea of this framework is to utilize large language models (LLMs) to provide explanations for comparatively lighter-weight pretrained language models (PLMs). Nevertheless, the proposed framework is of a universal nature and should be compatible with any PLMs. Investigations utilizing larger PLMs are reserved for future research endeavors.
Third, the process of prompting large language models to generate concept labels remains somewhat of an art. While we have proposed a systematic method for constructing desired prompts, the performance of the model may still be sensitive to the quality and structure of these prompts.
Lastly, while our proposed method shows promising results in English language tasks, it has not been tested extensively on other languages. This restricts its applicability in a multilingual setting. Future work should extend this method to other languages and conduct cross-lingual analysis.
We hope future research will build upon our work to address these limitations, moving us closer to truly interpretable, responsible, and universally applicable language models.

\section*{Ethics Statement}
In conducting this research, we strictly adhered to the \href{https://www.aclweb.org/portal/content/acl-code-ethics}{ACL Ethics Policy}. All data used in our work were either publicly available or anonymized, ensuring no personally identifiable information was involved. The work presented in this paper significantly contributes to the field of natural language processing and machine learning. By improving the interpretability of pre-trained language models, we are contributing to the creation of more transparent and trustworthy AI systems. This advancement is expected to have broad-ranging impacts across numerous domains that increasingly rely on AI, including healthcare, education, business, and finance, enhancing decision-making processes and user interaction with AI systems. However, the increased efficacy of these models could also raise potential societal concerns if not used responsibly. The misuse of these advanced NLP technologies could lead to privacy breaches, the propagation of misinformation, or the amplification of existing biases in data. As with any powerful technology, it is essential to consider its ethical implications and manage its deployment with care to ensure it's used for the betterment of society. Our work also underscores the need for continual research into strategies that mitigate potential bias in AI systems and protect user privacy. As researchers, we are committed to working towards these goals and urge those employing this technology to adhere to the same principles.



\bibliography{anthology,custom}
\balance
\bibliographystyle{acl_natbib}

\appendix
\nobalance
\section{Definitions of Training Strategies}
\label{app:def}

Given a text input $x \in \mathbb{R}^d$, concepts $c\in \mathbb{R}^k$ and its label $y$, the strategies for fine-tuning the text encoder $f_\theta$, the projector $p_\psi$ and the label predictor $g_\phi$ are defined as follows:

\noindent\textit{i) Vanilla fine-tuning a PLM:} The concept labels are ignored, and then the text encoder $f_\theta$ and the label predictor $g_\phi$ are fine-tuned either as follows:
\begin{equation*}
    \theta, \phi = \argmin_{\theta, \phi} L_{CE} (g_\phi(f_\theta(x), y),
\end{equation*}
or as follows (frozen text encoder $f_\theta$):
\begin{equation*}
    \phi = \argmin_{\phi} L_{CE} (g_\phi(f_\theta(x), y),
\end{equation*}
where $L_{CE}$ indicates the cross-entropy loss. In this work we only consider the former option for its significant better performance.

\noindent\textit{ii) Independently training PLM with the concept and task labels:} The text encoder $f_\theta$, the projector $p_\psi$ and the label predictor $g_\phi$ are trained seperately with ground truth concepts labels and task labels as follows:
\begin{equation*}
    \begin{aligned}
    \theta, \psi &= \argmin_{\theta, \psi} L_{CE} (p_\psi(f_\theta(x)),c), \\
    \phi &= \argmin_{\phi} L_{CE} (g_{\phi}(c),y).
    \end{aligned}
\end{equation*} 
During inference, the label predictor will use the output from the projector rather than the ground-truth concepts.

\noindent\textit{iii) Sequentilally training PLM with the concept and task labels:} We first learn the concept encoder as the independent training strategy above, and then use its output to train the label predictor:
\begin{equation*}
    \begin{aligned}
    \phi = \argmin_{\phi} L_{CE} (g_{\phi}(p_\psi(f_\theta(x),y).
    \end{aligned}
\end{equation*} 

\noindent\textit{iv) Jointly training PLM with the concept and task labels:} Learn the concept encoder and label predictor via a weighted sum $L_{joint}$ of the two objectives described above:
\begin{equation*}
\begin{aligned}
    \theta, \psi, \phi &= \argmin_{\theta, \psi, \phi} L_{joint}(x, c, y) \\ &= \argmin_{\theta, \psi, \phi} [L_{CE} (g_{\phi}(p_\psi(f_\theta(x),y) \\ &+ \gamma L_{CE} (p_\psi(f_\theta(x)),c)].
\end{aligned}
\end{equation*} 
 It's worth noting that the CBE-PLMs trained jointly are sensitive to the loss weight $\gamma$. We report the most effective results here, tested value for $\gamma$ are given in Table~\ref{tab:symbols} in Appendix~\ref{app:para}.


\section{Details of the Manual Concept Annotation for the IMDB Dataset}
\label{app:anno}

Our annotation policy is following a previous work~\cite{cai2021aspect} for NLP datasets annotating. For the \texttt{IMDB} dataset, we annotate the four concepts (Acting, Stroyline, Emotional Arousal, Cinematography) manually. Even though the concepts are naturally understandable by humans, two Master students familiar with sentiment analysis are selected as annotators for independent
annotation with the annotation tool introduced by \citet{yang2017yedda}. The strict quadruple matching F1 score between two annotators is $85.74\%$, which indicates a consistent agreement between the two annotators~\cite{kim2018feels}. In case of disagreement, a third expert will be asked to make
the final decision.

\section{Implementation Detail}
\label{app:implement}

In this section, we provide more details on the implementation settings of our experiments. Specifically, we implement our framework with PyTorch~\cite{paszke2017automatic} and HuggingFace~\cite{wolf2020huggingfaces} and train our framework on a single 80GB Nvidia A100 GPU. We follow a prior work~\citep{abraham2022cebab} for backbone implementation. All backbone models have a maximum token number of 512 and a batch size of 8. We use the Adam optimizer to update the backbone, projector, and label predictor according to Section~\ref{sec:setup}. The values of other hyperparameters (Table~\ref{tab:symbols} in Appendix~\ref{app:para}) for each specific PLM type are determined through grid search. We run all the experiments on an Nvidia A100 GPU with 80GB RAM.

\section{Parameters and Notations}
\label{app:para}

In this section, we provide used notations in this paper along with their descriptions for comprehensive understanding. We also list their experimented values and optimal ones, as shown in Table~\ref{tab:app_para}.
\begin{table*}[htbp]
\small
\setlength\tabcolsep{5pt}
\caption{Key parameters in this paper with their annotations and evaluated values. Note that \textbf{bold} values indicate the optimal ones.\label{tab:app_para}} 
\label{tab:symbols}
\centering
\begin{tabular}{@{}cccc@{}}
\toprule
\textbf{Notations}    & \textbf{Specification} & \textbf{Definitions or Descriptions}              & \textbf{Values}                  \\ \midrule
max\_len              & -                      & maximum token number of input                     & 128 / 256 / \textbf{512}                             \\
batch\_size           & -                      & batch size                                        & 8                                \\
plm\_epoch            & -                      & maximum training epochs for PLM and Projector     & 20                               \\
clf\_epoch            & -                      & maximum training epochs for the linear classifier & 20                               \\
hidden\_dim           & -                      & hidden dimension size                             & 128                              \\
emb\_dim              & LSTM                   & embedding dimension for LSTM                      & 300                              \\
\multirow{4}{*}{lr}   & LSTM                   & learning rate when the backbone is LSTM               & 1e-1 / \textbf{1e-2} / 5e-2 / 1e-3 / 1e-4 \\
                      & GPT2                   & learning rate when the backbone is GPT2               & 1e-3 / 5e-3/ \textbf{1e-4} / 5e-4/ 1e-5   \\
                      & BERT                   & learning rate when the backbone is BERT               & 1e-4 / 5e-4/ \textbf{1e-5} / 3e-5/ 5e-5   \\
                      & RoBERTa                & learning rate when the backbone is RoBERTa            & 1e-4 / 5e-4/ \textbf{1e-5} / 3e-5/ 5e-5   \\
\textbackslash{}gamma & -                      & loss weight in the joint loss $L_{joint}$                    & 0.1 / 0.3 / \textbf{0.5} / 0.7 / 1.0      \\
\textbackslash{}tau   & -                      & loss weight in the joint-MixUp loss $L_{jointMixUp}$              & 0.1 / 0.5 / \textbf{1.0} / 1.5 / 2.0      \\ \bottomrule
\end{tabular}
\end{table*}

\section{Statistics of Data Splits}
\label{app:split}
The Statistics and split policies of the experimented datasets, including the source concept dataset $\mathcal{D}_s$, the unlabeled concept dataset $\mathcal{D}_u$, and their augmented versions. The specific details are presented in Table~\ref{tab:stats_data}.

\begin{table*}[htbp]
\centering
\caption{Statistics of experimented datasets. $k$ denotes the number of concepts. \label{tab:stats_data}}
\scalebox{0.80}{
\begin{tabular}{@{}ccccc|cccc|c@{}}
\toprule
\multirow{2}{*}{\textbf{Dataset}} & \multicolumn{2}{c}{$\mathcal{D}_s$} & \multicolumn{2}{c|}{$\mathcal{D}_u$} & \multicolumn{2}{c}{$\Tilde{\mathcal{D}}_{sa}$} & \multicolumn{2}{c|}{$\Tilde{\mathcal{D}}_{u}$} & \multirow{2}{*}{\textbf{Task}} \\ \cmidrule(lr){2-9}
                                  & Train/Dev/Test    & $k$    & Train/Dev/Test     & $k$    & Train/Dev/Test    & $k$     & Train/Dev/Test    & $k$    &                                \\ \midrule
\texttt{CEBaB}                             & 1755/1673/1685    & 4    & 2000/500/500       & 0    & 1755/1673/1685    & 10    & 2000/500/500      & 10    & 5-way classification           \\ \midrule
\texttt{IMDB}                              & 100/50/50         & 4    & 1000/1000/1000     & 0    & 100/50/50         & 8     & 1000/1000/1000    & 8     & 2-way classification           \\ \bottomrule
\end{tabular}}
\end{table*}

\section{Statistics of Human-Annotated Concepts}
\label{app:stat_hac}

The Statistics of Human-Annotated Concepts in both \texttt{CEBaB} and \texttt{IMDB} datasets. We also include the accuracy of ChatGPT's concept prediction here. The specific details are presented in Table~\ref{tab:stats_concept_s}.

\begin{table*}[htbp]
\begin{center}
\caption{Statistics of human-specified concepts in $\mathcal{D}_s$ and the accuracy of ChatGPT's concept prediction. \label{tab:stats_concept_s}}
\scalebox{0.9}{
\begin{tabular}{@{}ccccccc@{}}
\toprule
\textbf{Dataset} ($\mathcal{D}_s$)       & \textbf{Concept}  & Negative      & Positive      & Unknown      & Total & \textbf{ChatGPT Acc.} \\ \midrule
\multirow{4}{*}{\texttt{CEBaB}} &Food&1693 (33.1\%)&2087 (40.8\%)&1333 (26.1\%)&5113&77.9\%\\
&Ambiance&787 (15.4\%)&994 (19.4\%)&3332 (65.2\%)&5113&69.2\%\\
&Service&1249 (24.4\%)&1397 (27.3\%)&2467 (48.2\%)&5113&78.7\%\\
&Noise&645 (12.6\%)&442 (8.6\%)&4026 (78.7\%)&5113&77.7\%

 \\ \midrule

\multirow{4}{*}{\texttt{IMDB}}  & Acting            & 76 (38\%)     & 66 (33\%)     & 58 (29\%)    & 200   & 73.0\%                \\
                       & Storyline         & 80 (40\%)     & 77 (38.5\%)   & 43 (21.5\%)    & 200   & 64.0\%                \\
                       & Emotional Arousal & 74 (37\%)     & 73 (36.5\%)   & 53 (26.5\%)  & 200   & 60.5\%                \\
                       & Cinematography    & 118 (59\%)    & 43 (21.5\%)   & 39 (19.4\%)  & 200   & 66.5\%                \\ \bottomrule
\end{tabular}}
\end{center}
\end{table*}

\section{Statistics of Concepts in Transformed Datasets}
\label{app:trans_con}

The Statistics and split policies of the transformed datasets of experimented datasets are presented in Table~\ref{tab:trans_con}.

\begin{table*}[htbp]
\begin{center}
\caption{Statistics of concepts in transformed datasets ($\Tilde{\mathcal{D}}$). Human-specified concepts are \underline{underlined}. Concepts shown in gray are not used in experiments as the portion of the "Unknown" label is too large.\label{tab:trans_con}}
\scalebox{0.9}{
\begin{tabular}{@{}cccccc@{}}
\toprule
\textbf{Dataset}      & \textbf{Concept}  & Negative      & Positive      & Unknown      & Total  \\ \midrule
\multirow{13}{*}{\texttt{CEBaB}} &\underline{Food}&2043(25.2\%)&4382(54.0\%)&1688(20.8\%)&8113\\
&\underline{Ambiance}&868(10.7\%)&1659(20.4\%)&5586(68.9\%)&8113\\
&\underline{Service}&1543(19.0\%)&2481(30.6\%)&4089(50.4\%)&8113\\
&\underline{Noise}&668(8.2\%)&477(5.9\%)&6968(85.9\%)&8113\\
&Cleanliness&55(0.7\%)&610(7.5\%)&7448(91.8\%)&8113\\
&Price&714(8.8\%)&527(6.5\%)&6872(84.7\%)&8113\\
&Location&303(3.7\%)&2598(32.0\%)&5212(64.2\%)&8113\\
&Menu Variety&238(2.9\%)&2501(30.8\%)&5374(66.2\%)&8113\\
&Waiting Time&572(7.1\%)&608(7.5\%)&6933(85.5\%)&8113\\
&Waiting Area&267(3.3\%)&1136(14.0\%)&6710(82.7\%)&8113\\
\rowcolor{gray!25}&Parking&53(0.7\%)&107(1.3\%)&7953(98.0\%)&8113\\
\rowcolor{gray!25}&Wi-Fi&9(0.1\%)&39(0.5\%)&8065(99.4\%)&8113\\
\rowcolor{gray!25}&Kids-Friendly&15(0.2\%)&536(6.6\%)&7562(93.2\%)&8113
 \\ \midrule                       
\multirow{9}{*}{\texttt{IMDB}}  
&\underline{Sentiment}&1624(50.7\%)&1576(49.2\%)&0(0.0\%)&3200\\
&\underline{Acting}&663(20.7\%)&1200(37.5\%)&1337(41.8\%)&3200\\
&\underline{Storyline}&1287(40.2\%)&1223(38.2\%)&690(21.6\%)&3200\\
&\underline{Emotiona Arousal}&1109(34.7\%)&1136(35.5\%)&955(29.8\%)&3200\\
&Cinematography&165(5.2\%)&481(15.0\%)&2554(79.8\%)&3200\\
&Soundtrack&107(3.3\%)&316(9.9\%)&2777(86.8\%)&3200\\
&Directing&537(16.8\%)&850(26.6\%)&1813(56.7\%)&3200\\
&Background Setting&288(9.0\%)&581(18.2\%)&2331(72.8\%)&3200\\
\rowcolor{gray!25}&Editing&304(9.5\%)&240(7.5\%)&2656(83.0\%)&3200\\
\bottomrule
\end{tabular}}
\end{center}
\end{table*}


\section{More Results on Explanable Predictions}
\label{app:logits}

Case studies on explanable predictions for both \texttt{CEBaB} and \texttt{IMDB} datasets are given in Figure~\ref{fig:logit_1} and Figure~\ref{fig:logit_2} respectively.

\begin{figure*}[htbp]
  \centering\scalebox{1.09}{
  \includegraphics[width=\linewidth]{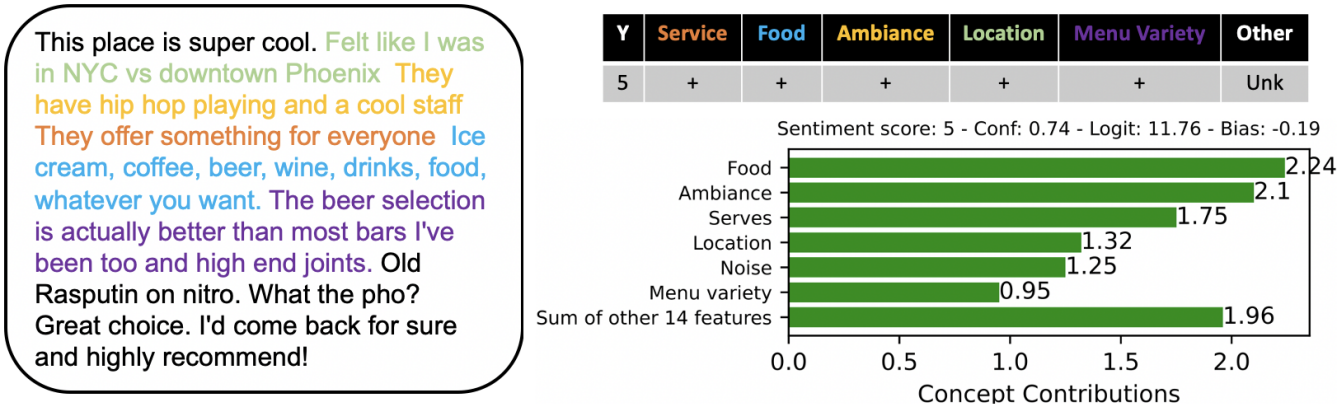}}
  \caption{{Illustration of the explanable prediction for an example from the \texttt{CEBaB} dataset.}}
  \label{fig:logit_1}
\end{figure*}

\begin{figure*}[htbp]
  \centering\scalebox{0.99}{
  \includegraphics[width=\linewidth]{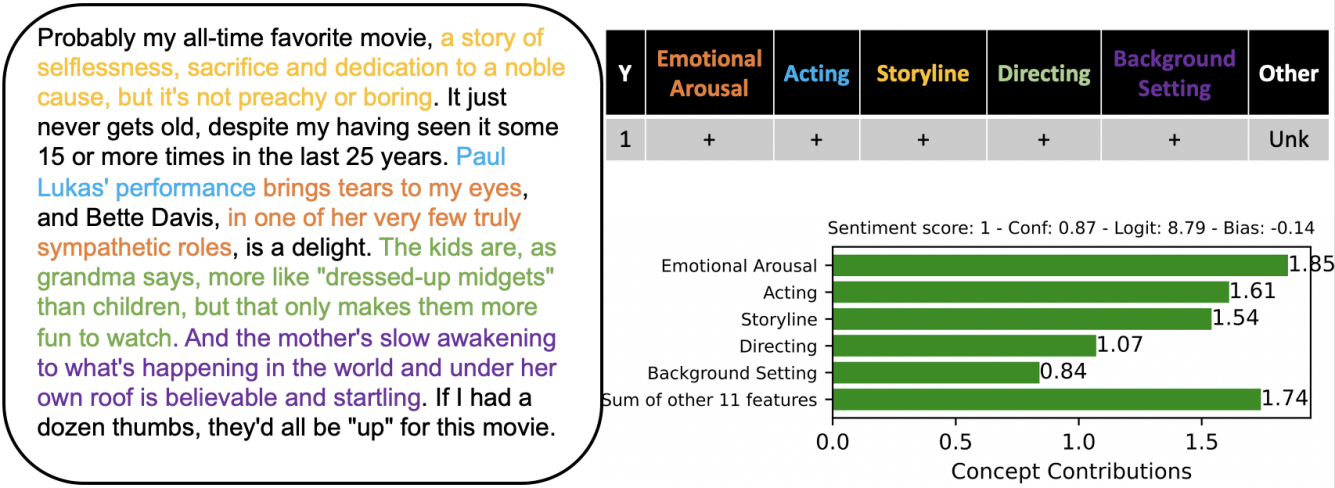}}
  \caption{{Illustration of the explanable prediction for an example from the \texttt{IMDB} dataset.}}
  \label{fig:logit_2}
\end{figure*}

\section{A case study on Test-time Intervention}
\label{app:internvene}

We present a case study of Test-time Intervention using an example from the transformed unlabeled concept data $\Tilde{\mathcal{D}}_u$ of the \texttt{CEBaB} dataset, as shown in Figure~\ref{fig:logit_3}. The first row displays the target concept labels generated by ChatGPT. The second row shows the predictions from the trained CBE-PLM model, which mispredicts two concepts ("Waiting time" and "Waiting area"). The third row demonstrates test-time intervention using ChatGPT as the oracle, which corrects the predicted task labels. Finally, the fourth row implements test-time intervention with a human oracle, rectifying the concept that ChatGPT originally mislabeled.

\begin{figure*}[htbp]
  \centering\scalebox{0.7}{
  \includegraphics[width=\linewidth]{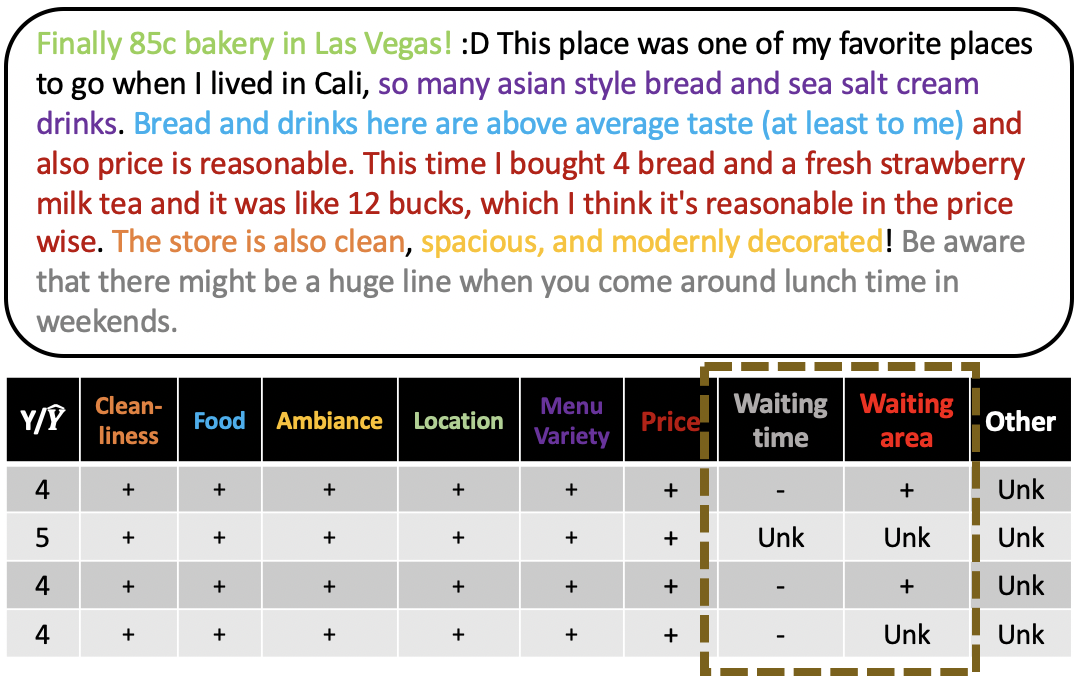}}
  \caption{{Illustration of the explanable prediction for an example from the transformed unlabeled concept data $\Tilde{\mathcal{D}}_{u}$ of the \texttt{CEBaB} dataset. The brown box with dash lines indicates the test-time intervention on corresponding concepts.}}
  \label{fig:logit_3}
\end{figure*}


\section{Examples of Querying ChatGPT}\label{app:gpt}

In this paper, we query ChatGPT for 1) augmenting the concept set, and 2) annotate missing concept labels. Note that in practice, we query ChatGPT (GPT4) via~\href{https://platform.openai.com/docs/api-reference/introduction}{OpenAI API}. Here we demonstrate examples from the ChatGPT (GPT4)~\href{https://chat.openai.com/?model=gpt-4}{GUI} for better illustration. The illustrations are given in Figure~\ref{fig:gpt_con} and Figure~\ref{fig:gpt_label}.

\begin{figure*}[htbp]
  \centering\scalebox{0.6}{
  \includegraphics[width=\linewidth]{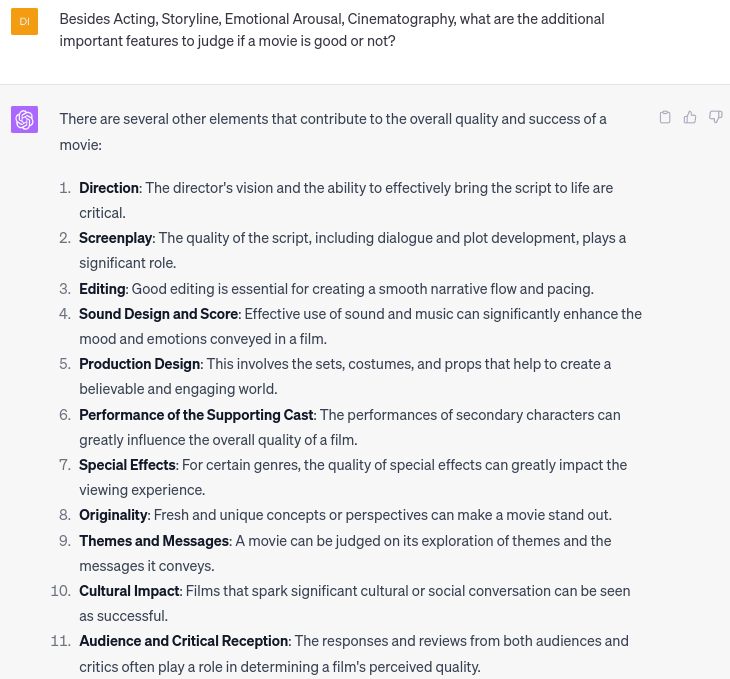}}
  \caption{{The illustration of querying ChatGPT for additional concepts for the \texttt{IMDB} dataset.}}
  \label{fig:gpt_con}
  \vspace{-0.1cm}
\end{figure*}

\begin{figure*}[htbp]
  \centering\scalebox{0.5}{
  \includegraphics[width=\linewidth]{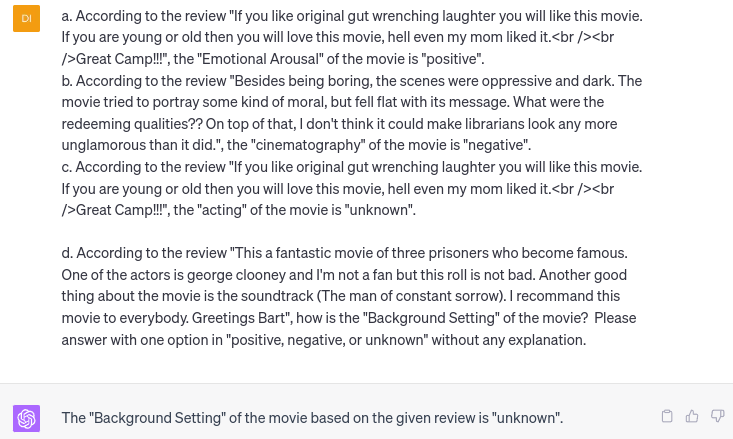}}
  \caption{{The illustration of querying ChatGPT for annotating a missing concept label for the \texttt{IMDB} dataset.}}
  \label{fig:gpt_label}
  \vspace{-0.1cm}
\end{figure*}

\end{document}